\documentclass{article} 
\usepackage{colm2024_conference}

\usepackage{microtype}
\usepackage{hyperref}
\usepackage{url}
\usepackage{booktabs}

\usepackage{amsmath}
\usepackage{amssymb}
\usepackage{mathtools}
\usepackage{amsthm}

\usepackage[capitalize,noabbrev]{cleveref}

\theoremstyle{plain}

\theoremstyle{definition}

\theoremstyle{remark}

\usepackage[textsize=tiny]{todonotes}

\usepackage{inconsolata}

\usepackage[utf8]{inputenc} 
\usepackage[T1]{fontenc}    
\usepackage{hyperref}       
\usepackage{xspace}
\usepackage{amsmath}
\usepackage{cleveref}
\crefname{section}{§}{§§}
\Crefname{section}{§}{§§}
\crefformat{subsubsection}{\S#2#1#3}
\usepackage{url}            
\usepackage{booktabs}       
\usepackage{amsfonts}       
\usepackage{nicefrac}       
\usepackage{microtype}      
\usepackage{xcolor}         
\usepackage{colortbl}
\usepackage{multirow}

\usepackage{appendix}
\usepackage{algpseudocode,algorithm,algorithmicx}
\usepackage{bm}
\usepackage{mathtools}
\usepackage{makecell}
\usepackage{enumitem}
\usepackage{subcaption}
\usepackage{lipsum}
\usepackage{wrapfig}
\usepackage{cuted}
\usepackage{xcolor}
\usepackage{adjustbox}
\usepackage{multirow}
\usepackage{graphicx}
\usepackage{gensymb}
\usepackage{listings}

\newcolumntype{H}{>{\setbox0=\hbox\bgroup}c<{\egroup}@{}}
\definecolor{cadmiumgreen}{rgb}{0.0, 0.52, 0.24}
\definecolor{mediumpersianblue}{rgb}{0.2, 0.4, 0.8}
\usepackage{pifont}
\definecolor{darkgreen}{rgb}{0,0.6,0}
\definecolor{codegray}{rgb}{0.5,0.5,0.5}
\definecolor{lightgray}{rgb}{0.7,0.7,0.7}
\definecolor{codepurple}{rgb}{0.07,0,0.53}
\definecolor{codered}{RGB}{189,41,0}
\definecolor{codecomment}{RGB}{153,153,153}
\definecolor{backcolour}{rgb}{0.96,0.96,0.96}
\definecolor{mygreen}{rgb}{0.0, 0.5, 0.0}
\definecolor{royalblue}{rgb}{0.0, 0.14, 0.4}
\definecolor{egyptianblue}{rgb}{0.06, 0.2, 0.65}
\definecolor{royalazure}{rgb}{0.0, 0.22, 0.66}
\definecolor{portlandorange}{rgb}{1.0, 0.35, 0.21}
\definecolor{saddlebrown}{RGB}{139,69,19}
\definecolor{sienna}{RGB}{183,105,68}
\definecolor{saddlebrown}{RGB}{139,69,19}
\hypersetup{
    colorlinks=true,
    linkcolor=sienna,
     urlcolor=royalblue,
    citecolor=egyptianblue,
}

\newcolumntype{H}{>{\setbox0=\hbox\bgroup}c<{\egroup}@{}}

\title{Dense Training, Sparse Inference: Rethinking Training of Mixture-of-Experts Language Models}

\author{Bowen Pan$^\diamond$ \hspace{3mm} Yikang Shen$^\dagger$ \hspace{3mm}  Haokun Liu$^\star$ \hspace{3mm} Mayank Mishra$^\dagger$ \hspace{3mm} Gaoyuan Zhang$^\dagger$
\\
\textbf{Aude Oliva$^{\diamond\dagger}$ \hspace{5mm} Colin Raffel$^\star$$^\ddagger$ \hspace{5mm} Rameswar Panda$^\dagger$}
\\
$^\diamond$MIT CSAIL, $^\dagger$MIT-IBM Watson AI Lab, $^\star$University of Toronto, $^\ddagger$Vector Institute
\\
\small{\texttt{\{bpan, oliva\}@mit.edu,}} \small{\texttt{\{haokunliu412, craffel\}@gmail.com,}} \\
\small{\texttt{\{rpanda, yikang.shen, mayank.mishra2, Gaoyuan.zhang\}@ibm.com}}
}

\colmfinalcopy 
\begin{document}

\maketitle

\begin{abstract}
Mixture-of-Experts (MoE) language models can reduce computational costs by 2-4$\times$ compared to dense models without sacrificing performance, making them more efficient in computation-bounded scenarios. However, MoE models generally require 2-4$\times$ times more parameters to achieve comparable performance to a dense model, which incurs larger GPU memory requirements and makes MoE models less efficient in I/O-bounded scenarios like autoregressive generation. In this work, we propose a hybrid dense training and sparse inference framework for MoE models (DS-MoE) which achieves strong computation and parameter efficiency by employing dense computation across all experts during training and sparse computation during inference. Our experiments on training LLMs demonstrate that our DS-MoE models are more parameter-efficient than standard sparse MoEs and are on par with dense models in terms of total parameter size and performance while being computationally cheaper (activating 30-40\% of the model's parameters). Performance tests using vLLM show that our DS-MoE-6B model runs up to $1.86\times$ faster than similar dense models like Mistral-7B, and between $1.50\times$ and $1.71\times$ faster than comparable MoEs, such as DeepSeekMoE-16B and Qwen1.5-MoE-A2.7B.
\end{abstract}

\section{Introduction}
While scaling up Large Language Models (LLMs) has proven to be an effective way to improve performance on a huge range of tasks, increased scale leads to increased computational costs. The Mixture-of-Experts (MoE) approach \cite{shazeer2017outrageously, fedus2022switch, shen2023moduleformer, jiang2024mixtral} presents one possible solution by selectively utilizing a subset of parameters for improved computational efficiency while maintaining or even enhancing performance. This efficiency is particularly beneficial in computation-bound scenarios where many tokens need to be processed simultaneously, like preprocessing a large batch of prompts. Despite these benefits, MoE models often require 2-4$\times$ more parameters than dense models \cite{dai2024deepseekmoe, shen2023moduleformer} to achieve comparable performance. The large number of parameters makes MoE models consume more memory and less efficient in I/O-bounded scenarios such as recurrent token generation. We hypothesize that the relative parameter inefficiency of MoE models is primarily due to the sparse training approaches typically used to train MoE models, wherein only a subset of experts is activated and optimized for each token. In addition, sparse training can lead to inefficient GPU utilization when expert parallelism is used and expert usage is unbalanced \cite{gale2023megablocks}.

\begin{figure}[h]
\centering
\includegraphics[width=\linewidth]{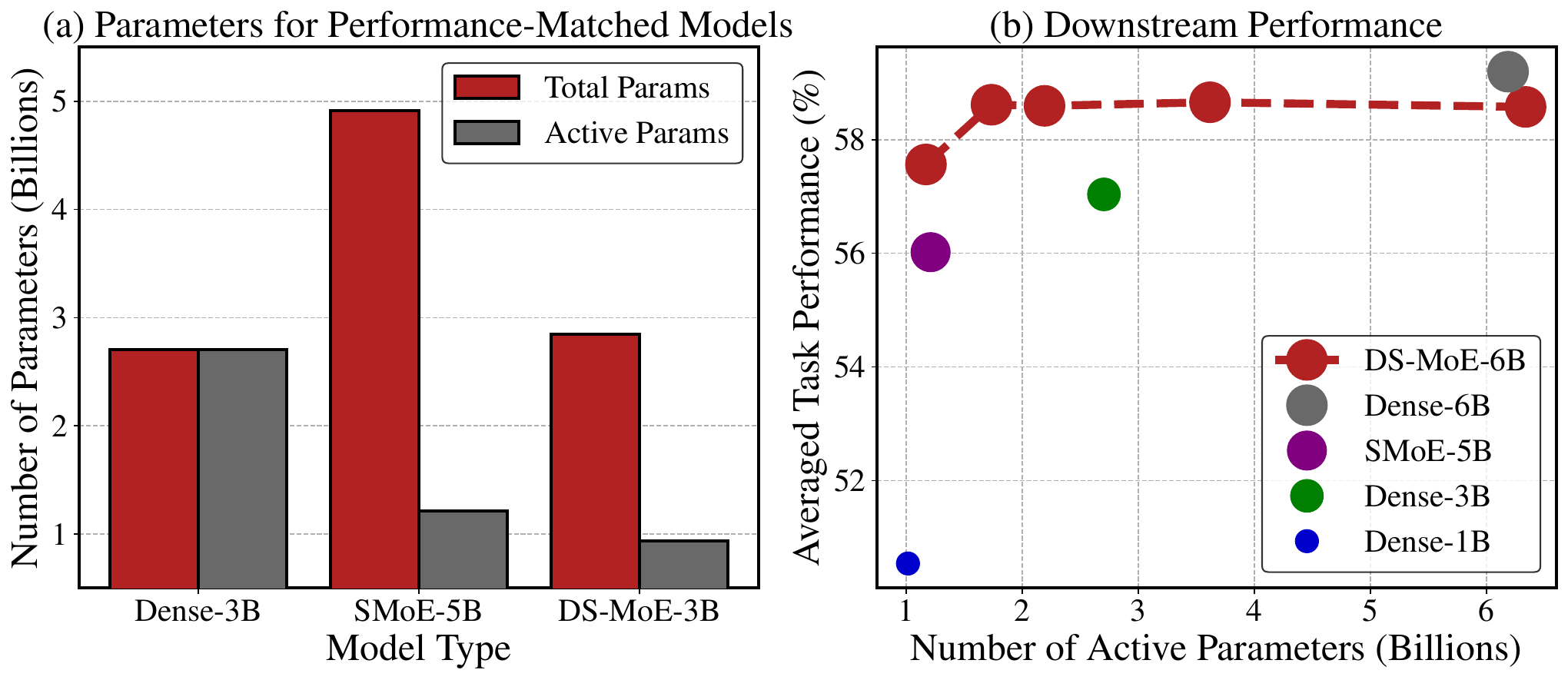}
\caption{Subfigure (a) showcases the sizes and computational profiles of the Dense-3B, SMoE-5B, and DS-MoE-3B models, each achieving a comparable averaged task performance in Table \ref{tab:base_performance}. The computational cost is quantified by counting the number of active parameters engaged during inference. Subfigure (b) displays the performance of our DS-MoE-6B model in sparse inference, set against that of the traditional dense models and SMoE models. The radius of the icon circle reflects the total number of the model parameters.}
\label{fig:teaser}
\end{figure}

In our study, we introduce dense training and sparse inference as a method to improve the parameter efficiency of MoE models. As illustrated in Figure \ref{fig:teaser}, our DS-MoE matches the performance of the same-size dense model while notably decreasing the number of active computing parameters during inference. In comparison with a performance-matched sparse MoE \cite{fedus2022switch, gale2023megablocks}, our DS-MoE significantly diminishes the total parameter count of the model while activating a similar number of parameters. The critical distinction between our DS-MoE and traditional sparse training lies in the involvement of \textbf{all experts} in each layer throughout the training phase. Additionally, we incorporate a Mutual Information (MI) loss \cite{shen2023moduleformer} that focuses on load balancing and expert concentration. This MI loss ensures the router produces an even distribution across experts and a sparse distribution for individual tokens, thereby ensuring expert use is balanced while allowing for sparse inference post-training. We then employ sparse inference by selecting the top \(K\) experts in each layer, based on their router scores. The value of \(K\) is either a fixed number or dynamically determined based on a predefined router score threshold \(\epsilon\). As a result, our DS-MoE model achieves performance comparable to that of dense models with same model size while only activating 30\% to 40\% of the parameters during inference.

Our experimental results demonstrate that: (1) Our DS-MoE significantly improves the parameter efficiency of MoE models and outperforms conventional sparse training methods for MoE; (2) when compared to  parameter-matched dense models, our DS-MoE model not only maintains comparable performance but also substantially reduces computation by activating 30-40\% of parameters during inference;  (3) we observe that larger models exhibit greater tolerance to sparsity, effectively maintaining dense-inference performance levels by engaging fewer experts, and (4) our DS-MoE has the best throughput performance in both computation-bounded and I/O-bounded scenarios.
\vspace{-1mm}
\section{Related Work}
\vspace{-1mm}
\paragraph{Spa60rsely Gated Mixture-of-Experts Models} Shazeer et al. \cite{shazeer2017outrageously} introduced the sparsely gated Mixture of Experts (MoE) layer, which are sparsely activated subnetworks whose routing is determined by softmax gates, to enhance the scalability of LSTMs \cite{Hochreiter1997LongSM}. With the advent of transformers \cite{Vaswani2017AttentionIA, Radford2018ImprovingLU, Devlin2019BERTPO}, which have shown significant improvements with scaling \cite{kaplan2020scaling}, MoE has been recognized as a promising avenue for advancing model performance. The integration of MoE into the Transformer framework has been furthered by innovations such as Gshard \cite{Lepikhin2020GShardSG}, Switch transformers \cite{fedus2022switch}, and GLAM \cite{Du2021GLaMES}, which each introduce parallelization techniques for efficient training of MoE models across multiple devices. Nonetheless, the complexity of training MoE models is augmented by the routing problem \cite{Rosenbaum2019RoutingNA, Mittal2022IsAM}, prompting numerous studies to enhance routing strategies either by redesigning the router \cite{Roller2021HashLF, Lewis2021BASELS, Chi2022OnTR, Zhou2022MixtureofExpertsWE} or by altering the training methodology \cite{Zoph2022STMoEDS, Dai2022StableMoESR, Shen2023ModuleFormerLM}. Additional research has explored converting dense models into MoE models \cite{Komatsuzaki2022SparseUT, Zhang2021MoEficationTF} or utilizing a complete transformer model as an expert \cite{Li2022BranchTrainMergeEP}. Beyond merely scaling, MoE offers benefits in managing diverse tasks, with notable achievements in machine translation \cite{Kudugunta2021BeyondDT}, multitask learning \cite{Hazimeh2021DSelectkDS, Gupta2022SparselyAM}, and instruction tuning \cite{Shen2023MixtureofExpertsMI}. The landscape of MoE models is rapidly expanding, with recent introductions of more powerful frameworks \cite{jiang2024mixtral, dai2024deepseekmoe}, and studies by \cite{shen2023mixture} and \cite{qiu2023emergent} have shown that instruction finetuning significantly enhances MoE models, bolstering their prevalence.

\paragraph{Sparsity in Dense Model.}Sparsity is a common trait in large language models. \cite{liu2023deja} has demonstrated that similar performance levels can be achieved by activating only 10-20\% of the neurons in these models. The concept of MoEfication, as introduced by another study \cite{zhang2021moefication}, involves organizing neurons into distinct expert groups within a dense model and then converting it into a sparse MoE model. This transformation is accomplished by training a router to manage these expert groups, thereby preserving performance. Model pruning techniques \cite{voita2019analyzing,michel2019sixteen} leverage the inherent sparsity within dense models to eliminate superfluous neurons, enhancing efficiency. Similarly, dynamic inference strategies \cite{wang2018skipnet,wu2018blockdrop,pan2021va} aim to selectively engage only the necessary parts of the model during computation, optimizing resource use. Furthermore, studies on model-preserving compression \cite{chee2022model} have revealed that neurons in dense models are often redundant. It has been shown that the model's parameter size can be significantly reduced by generating portions of the neuron through cost-effective operations on existing neurons, further affirming the potential for optimization in model design.

\paragraph{Efficient Inference for LLM.} The high computational costs of large-scale LLMs have led to a great deal of work that aims to make inference of LLMs more efficient. Structured pruning techniques \cite{xia2023sheared,xia2022structured,cai2019once,wen2016learning,liu2017learning,luo2017thinet} aim to trim a pre-trained large model in a systematic manner, resulting in a smaller model that can be further enhanced through continuous learning. Quantization methods \cite{xiao2023smoothquant,nagel2019data,nagel2020up,wang2019haq,lin2023awq,frantar2022gptq} significantly reduce the model size and notably accelerate inference speed. Additionally, speculative decoding strategies \cite{stern2018blockwise,chen2023accelerating,leviathan2023fast} expedite LLM inference by employing a compact draft model to decode tokens concurrently, showcasing innovative approaches to improve computational efficiency and model performance.

\vspace{-2mm}
\section{Methods}
\vspace{-2mm}
\begin{figure*}
\centering
\includegraphics[width=\linewidth]{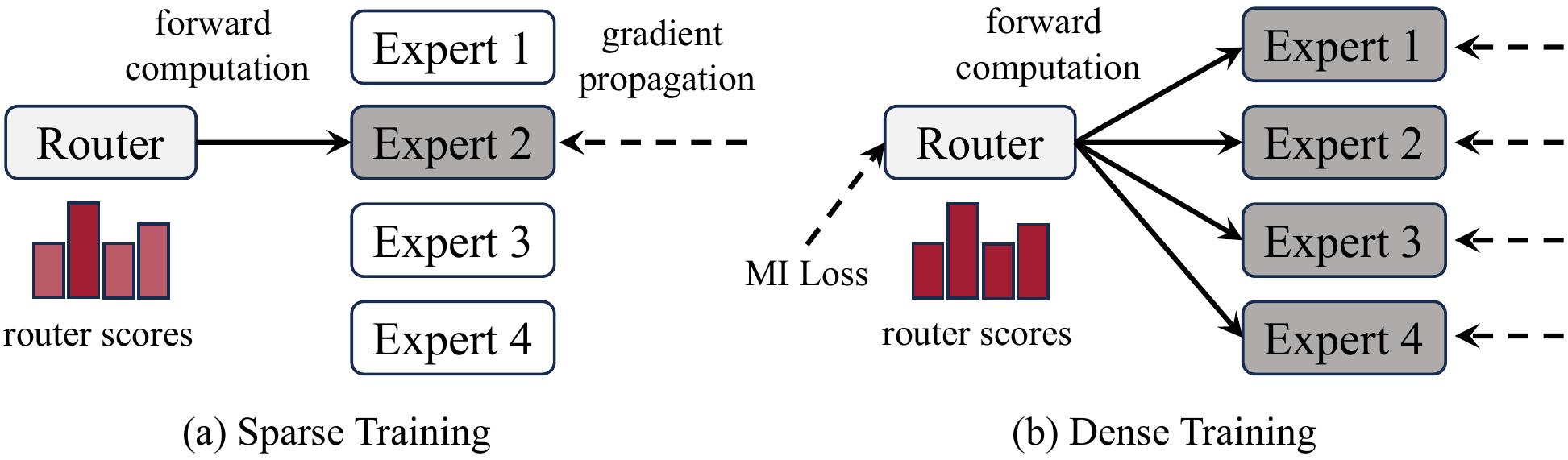}
\caption{Illustration of Dense Training of MoE models: Subfigure (a) illustrates the conventional sparse training method in MoE models, characterized by sparse gradient propagation in both the router and the experts. In subfigure (b), we detail the dense training strategy in our DS-MoE, which involves dense propagation of gradients for both routers and experts.}
\label{fig:method}
\end{figure*}

In this section, we first provide an overview of the MoE language model. Subsequently, we present our DS-MoE framework, detailing the process of densely training our MoE model prior to sparse inference, as well as two pivotal elements for DS-MoE framework: the mutual information (MI) loss and the Mixture of Attention Head (MoA) blocks.

\vspace{-2mm}
\subsection{Preliminary: Mixture-of-Experts Language Models} \label{sec:preliminary}
We take the feed-forward network (FFN) in the transformer language model for example to illustrate the MoE architecture. An MoE FFN comprises $N$ experts and a router $\mathbf{h}$, where each expert $\mathbf{e}$ is an MLP module and the router predicts a score for each expert. Given an input token $\mathbf{X}\in \mathbb{R}^{d_h}$, the MoE computes the output $\mathbf{O}\in \mathbb{R}^{d_h}$ through:
\begin{align}
     \mathbf{S} = \mathrm{softmax}\big(\mathbf{h(X)}\big), \quad
     \mathbf{O} = \sum_{i=1}^{K}S_{\mathbf{A}_i}\cdot\mathbf{e}_{\mathbf{A}_i}\mathbf{(X)},
\end{align}
where $\mathbf{S}\in \mathbf{R}^N$ is the score vector for the experts and $\mathbf{A}_i$ is the index for $i^{th}$ of the $K$ experts with the highest scores. During training, gradients only backpropagate through the selected experts $\mathbf{e}_{topK_i}$ and the corresponding scores $S_{topK_i}$. 

\subsection{DS-MoE Framework}
The traditional MoE language models, despite being able to match the performance of dense models with approximately 40-50\% of the computation, necessitate significantly more parameters, typically 2-3 times more. This increased requirement for parameters arises from the process of backpropagation in MoE models, which is sparsely optimized as we discussed in Section \ref{sec:preliminary}, thus not as efficient as in dense models. Our approach trains the MoE model in a dense manner with an additional MI loss, and performs inference sparsely. This strategy aims to retain the runtime efficiency of traditional MoE models while achieving the parameter efficiency of dense models.

\paragraph{Dense Training.} 
The fundamental concept of dense training revolves around optimizing the router using complete gradients. Unlike traditional MoEs, where the gradient of \(\mathbf{S}\) is expressed as:
\begin{align}
    \mathbf{\nabla S} = \left[\mathbf{e}_1(\mathbf{X}),..., \mathbf{e}_N(\mathbf{X})\right]^\intercal \mathbf{\nabla O} \odot \mathbf{M},
\end{align}
in this context, \(\mathbf{M}\in \{0,1\}^{N}\) serves as a binary mask identifying the activated experts. Specifically, \(\mathbf{M}_{i}=1\) indicates that expert \(\mathbf{e}_i\) is active in the forward pass, and \(\mathbf{M}_{i}=0\) otherwise. To preserve all gradients of \(\mathbf{S}\), we ensure that the output of every expert is computed during the forward pass, and this output is retained for use during backpropagation. The gradients of \(\mathbf{S}\) and \(\mathbf{e}_i(\mathbf{X})\) in our approach are articulated as follows:
\begin{align}
    & \mathbf{\nabla S} = \left[\mathbf{e}_1(\mathbf{X}),..., \mathbf{e}_N(\mathbf{X})\right]^\intercal \mathbf{\nabla O}, \quad
     \mathbf{\nabla e}_j\mathbf{(X)} = S_j\nabla \mathbf{O},
\end{align}
where $j$ represents the expert index. As shown in Figure \ref{fig:method}, our approach to training the MoE model densely involves activating all experts. 

\paragraph{Sparse Inference.}
During inference, only the top \(K\) experts, determined by their scores, are used. The selection of \(K\) is based either on a predetermined value or adaptively, depending on how many experts have scores above a specified threshold \(\epsilon\). We adopt the SimpleMoE~\cite{tan2024scattered} implementation for the sparse inference. More details can be found in Section \ref{sec:inference}.

\paragraph{Mutual Information Loss.} 
To achieve load balance among all experts and prevent the underutilization of model capacity, we integrate a Mutual Information (MI) loss into the router. This method, following \cite{shen2023moduleformer}, aims to maximize the entropy of the expert distribution to ensure even distribution of workload:
\begin{align}
    H(\mathbf{e}) = - \sum_{i=1}^{N}p(\mathbf{e})\log p(\mathbf{e}),
\end{align}
thereby promoting load balance across experts. In addition, to avoid the router adopting overly simplistic solutions and to ensure expert concentration, we minimize the conditional entropy of the expert distribution, \( H(\mathbf{e}|\mathbf{X}) \). The overall loss function is therefore defined as:
\begin{align}
    \mathcal{L}_{\mathrm{MI}} & = -  H(\mathbf{e}) + \frac{1}{|\mathcal{X}|} \sum_{\mathbf{X}\in \mathcal{X}} H(\mathbf{e}|\mathbf{X}),
\end{align}
where \( \mathcal{X} \) denotes the tokens in a minibatch. This approach not only ensures an equitable load balance among the experts but also maintains a high level of concentration on the appropriate solutions, optimizing the router's performance. The total loss is then calculates as
\begin{align}
    \mathcal{L} & = \mathcal{L}_{\mathrm{LM}} + \alpha \mathcal{L}_{\mathrm{MI}}
    \label{eq:total-loss}
\end{align}
where $\mathcal{L}_{\mathrm{LM}}$ is the standard autoregressive language modeling loss and $\alpha$ is the weight for mutual information loss.

\paragraph{Mixture of Attention Head.} 
Unlike the majority of current MoE language models that retain a dense layer for self-attention, we have substituted our self-attention layer with a Mixture of Attention (MoA) heads layer \cite{zhang2022mixture}. Our MoA heads are constructed following the usage of group-query attention (GQA) \cite{ainslie2023gqa}, where key and value pairs are shared among a group of query vectors. In our implementation, each expert in the MoA heads is responsible for computing $N_{\mathrm{head}}$ query vectors $\mathbf{Q}\in \mathbb{R}^{d_\mathrm{{head}}}$. For a given input token $\mathbf{X}\in \mathbb{R}^{d_\mathrm{h}}$, the output from an expert $e_i$ is derived as follows:
\begin{align}
\mathbf{Q}_{i} = \mathbf{W_q}\mathbf{X}, \quad
    \mathbf{O}_{ij} = \mathrm{softmax}(\mathbf{Q}_{ij}\mathbf{K}_j^\intercal)\mathbf{V}_j\mathbf{W_o}_j,
\end{align}
here, $\mathbf{W_q}\in \mathbb{R}^{N_{\mathrm{head}}\times d_\mathrm{{head}}\times d_\mathrm{{h}}}$ represents the query projection weight for expert $\mathbf{e}_i$. It is important to note that the key and value cache, represented by $\mathbf{K}, \mathbf{V}\in \mathbb{R}^{N\mathrm{head} \times L\times d\mathrm{head}}$, where $L$ is the length of the cache, is shared among all the experts.
The output projection for the expert is indicated by $\mathbf{W_o}\in \mathbb{R}^{N_{\mathrm{head}}\times d_\mathrm{head}\times d_\mathrm{h}}$. The final output of the layer is calculated as:
\begin{align}
    \mathbf{O} = \sum_{k=1}^K S_{\mathbf{A}_k} \sum_{j=1}^{N_\mathrm{head}}\mathbf{O}_{\mathbf{A}_kj},
\end{align}
where $\mathbf{A}$ is the index set for the activated experts.

\section{Empirical Study}
In this section, we comprehensively evaluate our DS-MoE, focusing on its performance in downstream tasks, sparsity, and GPU inference speed. The primary objective of our study is to investigate the advantages of DS-MoE compared to both dense models and SMoE models. We test our model and baselines in moderate-scale language modeling.

\subsection{Experimental Setup}

\paragraph{Dataset and Tokenization}
We pretrain our models using a subset of the Pile dataset \cite{gao2020pile}, and apply tokenization using the CodeGen tokenizer \cite{nijkamp2022codegen}. This dataset encompasses 300B tokens. Specifically, we utilize a 30B token subset for training our 1B-scale models and a 100B token subset for the training of models at the 3B and 6B scales.

\begin{table*}[h]
    \centering
    \small
    \caption{Model Architecture Hyperparameters. Here, $N_\mathrm{att}$ and $N_\mathrm{ffd}$ denote the number of experts in the self-attention layer and the MLP layer, respectively. In the case of the SMoE models, the top-2 experts are activated both during training and inference phases.}
    \begin{tabular}{c|ccccccccc}
        \toprule
        \multirow{2}{*}{Model} & \multirow{2}{*}{$D_{\text{emb}}$} & \multirow{2}{*}{$N_\mathrm{layer}$} & \multirow{2}{*}{$N_\mathrm{att}$} & \multirow{2}{*}{$N_\mathrm{head}$} & \multirow{2}{*}{$D_\mathrm{att}$} & \multirow{2}{*}{$N_\mathrm{ffd}$} & \multirow{2}{*}{$D_\mathrm{ffd}$} & total \\
         & & & & & & & & params \\
        \midrule
        Dense-1B & 2048 & 24 & 1 & 32 & 64 & 1 & 8192 & 1017M  \\
        Dense-3B & 3072 & 28 & 1 & 32 & 96 & 1 & 12288 & 2705M \\
        Dense-6B & 4096 & 36 & 1 & 32 & 128 & 1 & 16384 & 6186M \\
        \midrule
        SMoE-1B & 2048 & 24 & 1 & 32 & 64 & 8 & 1024 & 1042M \\
        SMoE-1.5B & 2048 & 24 & 1 & 32 & 64 & 12 & 1024 & 1445M  \\
        SMoE-5B & 3072 & 28 & 1 & 32 & 96 & 16 & 1536 & 4911M \\
        \midrule
        DS-MoE-1B & 2048 & 24 & 16 & 2 & 64 & 32 & 256 & 1067M \\
        DS-MoE-3B & 3072 & 28 & 8 & 4 & 96 & 32 & 384 & 2846M \\
        DS-MoE-6B & 4096 & 36 & 8 & 4 & 128 & 32 & 512 & 6343M  \\
        \bottomrule
    \end{tabular}
    \label{tab:model_arch}
\end{table*}

\paragraph{Model Hyperparameters.} 
We list the hyperparameter settings of different model architectures in Table~\ref{tab:model_arch}. Here, $N_\mathrm{att}$ and $N_\mathrm{ffd}$ represent the number of experts in each attention layer and each feed-forward layer respectively. In our models, we use the GeLU \cite{hendrycks2016gaussian} activation function. We use Grouped-Query Attention (GQA) \cite{ainslie2023gqa} in our attention blocks. We use 2 shared key-value heads for the 1B models and 4 for the 3B and 6B models. 

\begin{wraptable}{r}{0.45\textwidth}
    \centering
    \caption{Value of $\alpha$ in Our Models.}
    \begin{tabular}{c|cc}
        \toprule
        Model  & $\alpha$ in MoA & $\alpha$ in MoE \\
        \midrule
        DS-MoE-1B & $3.5e-4$ & $6.3e-4$ \\
        DS-MoE-3B & $2e-4$ & $4e-4$ \\
        DS-MoE-6B & $2e-4$ & $2e-4$ \\
        \bottomrule
    \end{tabular}
    \label{tab:alpha}
\end{wraptable}
\paragraph{Training Details.} We train our models using the AdamW optimizer \cite{loshchilov2017decoupled} with a learning rate of $3 \times 10^{-4}$. The training includes a cosine learning rate schedule with a warmup of 1 billion tokens for 1B models and 2 billion for 3B and 6B models. We use a constant weight decay of 0.01 and clip gradients at 1.0 throughout the training. Batch sizes are 0.5 million tokens for 1B models and 2 million for 3B and 6B models, with a sequence length of 2048 tokens. To optimize training, we use fully sharded data parallelism \cite{zhao2023pytorch, rajbhandari2020zero} and activation checkpointing \cite{korthikanti2023reducing}. Training times are 24 hours for 1B models on 8 H100 80GB GPUs, 64 hours for 3B models and 124 hours for 6B models on 32 H100 GPUs. The mutual information loss weights ($\alpha$ in Equation \ref{eq:total-loss}) are listed in Table \ref{tab:alpha}.  

\subsubsection{Evaluation Settings}
\paragraph{Baselines.} We compare our method against two baselines. \textit{A. Dense model.} For each instance of our DS-MoE model, we train an analogous dense model. This model is designed to have a parameter size similar to that of our DS-MoE model, as detailed in Table \ref{tab:model_arch}. The only parameter difference between the dense model and our DS-MoE model arises from the router function. \textit{B. Sparse MoE.} We train the MoE model with traditional sparse gradient propagation to match the performance of Dense-1B and Dense-3B models. The Top-K is set to be 2 with the use of switch loss \cite{fedus2022switch} for the load balance of the routers. For the implementation of the sparse MoE block, we employ dMoE \cite{gale2023megablocks}. The model hyperparameters for our SMoE baselines are outlined in Table \ref{tab:model_arch}. These baseline models aim to highlight the parameter inefficiencies found in traditional sparse MoE training approaches.

\paragraph{Downstream tasks.} We assess our models across a diverse array of downstream tasks, encompassing both common-sense reasoning and question-answering. These include evaluations on PiQA \cite{bisk2020piqa}, HellaSwag \cite{zellers2019hellaswag}, WinoGrande \cite{sakaguchi2021winogrande}, SciQ \cite{welbl2017crowdsourcing}, and Arc~\cite{allenai:arc}. Additionally, we measure and report the model's perplexity on the Wikitext dataset \cite{wikitext}. For all these evaluations, we utilize the LM evaluation harness \cite{eval-harness} to ensure consistency and reliability in our testing methodology.

\begin{table*}
    \centering
    \caption{Evaluation of Base Models in Zero-shot and Language Modeling Tasks. The number of active parameter and the percentage of the active hidden are calculated across all the downstream tasks and the wikitext dataset. Acronyms: HS (HellaSwag), WG (WinoGrande).
    }
    \begin{adjustbox}{width=\textwidth,center}
        \begin{tabular}{c|Hcccccc|cc|cc}
        \toprule
        \multirow{2}{*}{Model} & \multirow{2}{*}{Gradients} & \multirow{2}{*}{HS} & \multirow{2}{*}{PIQA} & \multirow{2}{*}{WG} & \multirow{2}{*}{SciQ} & \multirow{2}{*}{Arc-e} & \multirow{2}{*}{Arc-c} & Avg. & Wikitext & Active & Active\\
         & & & & & & & & Perf.$\uparrow$ & PPL$\downarrow$ & Params & Hidden \\
        \midrule
        Dense-1B  & Dense & 33.1 & 66.6 & 51.1 & 80.0 & 50.8 & 21.5 & 50.5 & 20.48 & 1017M & 100\% \\
        SMoE-1B & Sparse & 32.8 & 66.4 & 52.4 & 79.7 & 50.7 & 21.7 & 50.5 & 21.09 & 419M & 40\% \\
        SMoE-1.5B & Sparse & 33.1 & 67.7 & 52.5 & 79.7 & 50.5 & 22.8 & 51.0 & 20.32 & 419M & 29\% \\
        DS-MoE-1B & Dense & 33.7 & 68.1 & 50.8 & 81.1 & 52.4 & 22.2  & 51.4 & 20.37 & 439M & 41\%\\
        
        \midrule
        Dense-3B & Dense & 40.4 & 71.4 & 58.7 & 86.0 & 59.6 & 26.1 & 57.0 & 14.77 & 2705M & 100\%\\
        SMoE-5B & Sparse & 40.1 & 70.7 & 56.5 & 85.6 & 58.4 & 24.8 & 56.0 & 14.93 & 1212M & 25\% \\
        DS-MoE-3B & Dense & 39.3 & 71.6 & 57.9 & 85.6 & 57.7 & 24.9 & 56.2 & 15.48 & 934M & 34\% \\
        \midrule
        Dense-6B  & Dense & 44.3 & 72.2 & 59.9 & 88.0 & 62.9 & 27.9 & 59.2 & 12.98 & 6186M & 100\% \\
        DS-MoE-6B & Dense &  43.5	& 73.0 & 57.9 & 86.9 & 61.9 & 27.9 & 58.5 & 13.89 & 1813M & 29\% \\
        \bottomrule
    \end{tabular}
    \label{tab:base_performance}
    \end{adjustbox}
\end{table*}

\subsection{Results}\label{sec:exp-results}
In Table~\ref{tab:base_performance}, we count the mean active parameters across a range of zero-shot tasks as well as the Wikitext \cite{wikitext} language modeling task. Additionally, we evaluate the proportion of active parameters within the hidden layers. For DS-MoE-1B model, we evaluate the performance of the dense model baseline and the sparse training baseline. For all the DS-MoE models, experts are activated based on a criterion where their normalized probability
\footnote{The normalized probability is calculated by multiplying the router's output probability by the total number of experts.} 
exceeds a threshold $\epsilon$. The threshold $\epsilon$ can be flexibly adjusted to balance the performance and sparsity.

\begin{figure*}
\centering
\includegraphics[width=0.95\linewidth]{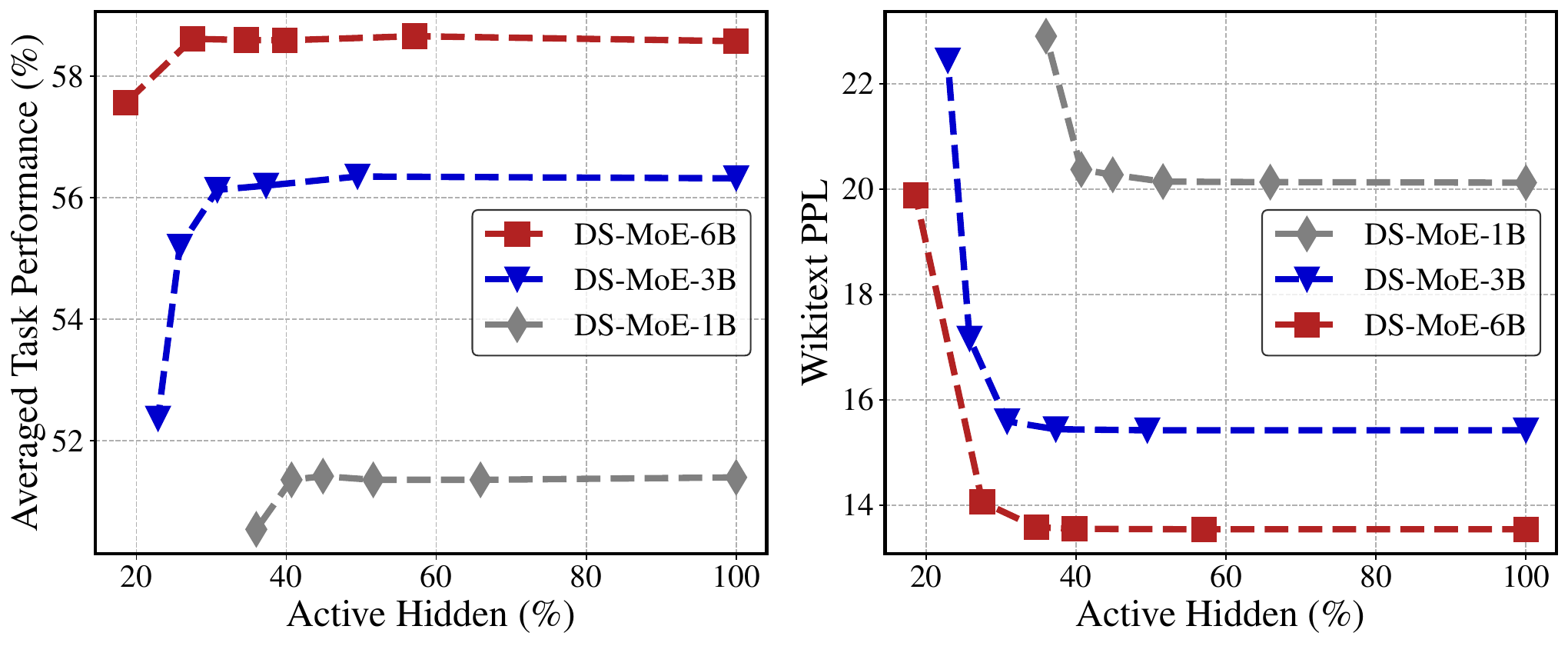}
\caption{We assess the sparsity in our DS-MoEs by gradually deactivating experts to attain increasingly sparse configurations, monitoring until a significant performance drop occurs.}
\label{fig:sparsity}
\end{figure*}
We list the evaluation results of the baselines and our models Table~\ref{tab:base_performance}, from which we derive three key insights.
\textbf{Firstly}, it is evident that the DS-MoE model demonstrates superior parameter efficiency compared to its sparsely trained counterpart.  For example, the table reveals that our DS-MoE-3B model not only aligns with the SMoE-5B model in terms of performance and computational expenses but does so with half the number of expert parameters in the MLP layers compared to the SMoE-5B model. This parameter efficiency improves the inference throughput when I/O is bounded as we demonstrated in Section \ref{sec:inference}.
\textbf{Secondly}, applying dense optimization to experts can achieve comparable parameter efficiency to that of traditional dense models. The table demonstrates that, across all three varying sizes, our DS-MoE models either closely match or even surpass their dense model counterparts in downstream task performance and language modeling capabilities. This model performance is achieved with a way lesser count of active parameters, thereby significantly reducing computational costs. \textbf{Thirdly}, the sparsity observed in DS-MoE models intensifies as the model size expands. This increase in sparsity is evident from the rising ratio of activated parameters within the hidden layers, indicating a clear trend across our DS-MoE models from 1B to 6B in size. Also as illustrated in Figure \ref{fig:sparsity}, there is a strategic reduction in the number of sampled experts to a point where further reduction noticeably degrades performance. This is visually represented by the turning point moving towards the left (indicating increased sparsity) as the model size grows.   This pattern suggests an even higher level of sparsity in models of a larger magnitude, including those with over 70B parameters.

\subsection{Ablation Study and Analysis}\label{sec:abl}

\begin{wraptable}{r}{0.6\textwidth}
    \centering
    \caption{Effect of Different $\alpha$ on our DS-MoE-6B model.}
    \begin{tabular}{cc|c|cc}
        \toprule
         Active & Active & \multirow{2}{*}{$\alpha_{\text{mlp}}$} & \multirow{2}{*}{Avg.} & Wikitext \\
         Params & Hidden & &  & PPL \\
        \midrule
        1826M & 29\% & $4e-4$ & 57.8 & \textbf{13.9} \\
        1813M & 29\% & $2e-4$ & \textbf{58.5} & \textbf{13.9} \\
        \midrule
        1496M & 24\% & $4e-4$ & \textbf{57.8} & \textbf{14.0} \\
        1497M & 24\% & $2e-4$ & 56.9 & 16.1 \\
        \bottomrule
    \end{tabular}
    \label{tab:6b-abl}
\end{wraptable}
\paragraph{Effect of $\alpha$.} To explore the impact of the mutual information loss weight on model sparsity and performance, we conduct an ablation study using our DS-MoE-6B models. In this study, we fix the $\alpha$ value at $2e-4$ in the self-attention layer, while in the MLP layer, we vary the weight from $2e-4$ to $4e-4$. This adjustment is made during the model training phase. For evaluation purposes, we modulate the $\alpha$ value to ensure that both models operated at identical sparsity levels. We assess the models' performance on zero-shot tasks and their Wikitext perplexity at two sparsity levels: 24\% and 29\%. According to the results presented in Table \ref{tab:6b-abl}, the model trained with $\alpha=4e-4$ demonstrates resilience at higher sparsity levels, maintaining its performance across both tested sparsity thresholds. Conversely, the model trained with $\alpha=2e-4$ exhibits diminished performance at the 24\% sparsity level, though it has the best performance at the 29\% sparsity level. Hence, we deduce that the $\alpha$ parameter plays a pivotal role in balancing the model's tolerance to high sparsity against its overall performance.

\begin{wrapfigure}{r}{0.52\textwidth}
  \centering
  \vspace{-3mm}
  \includegraphics[width=0.5\textwidth]{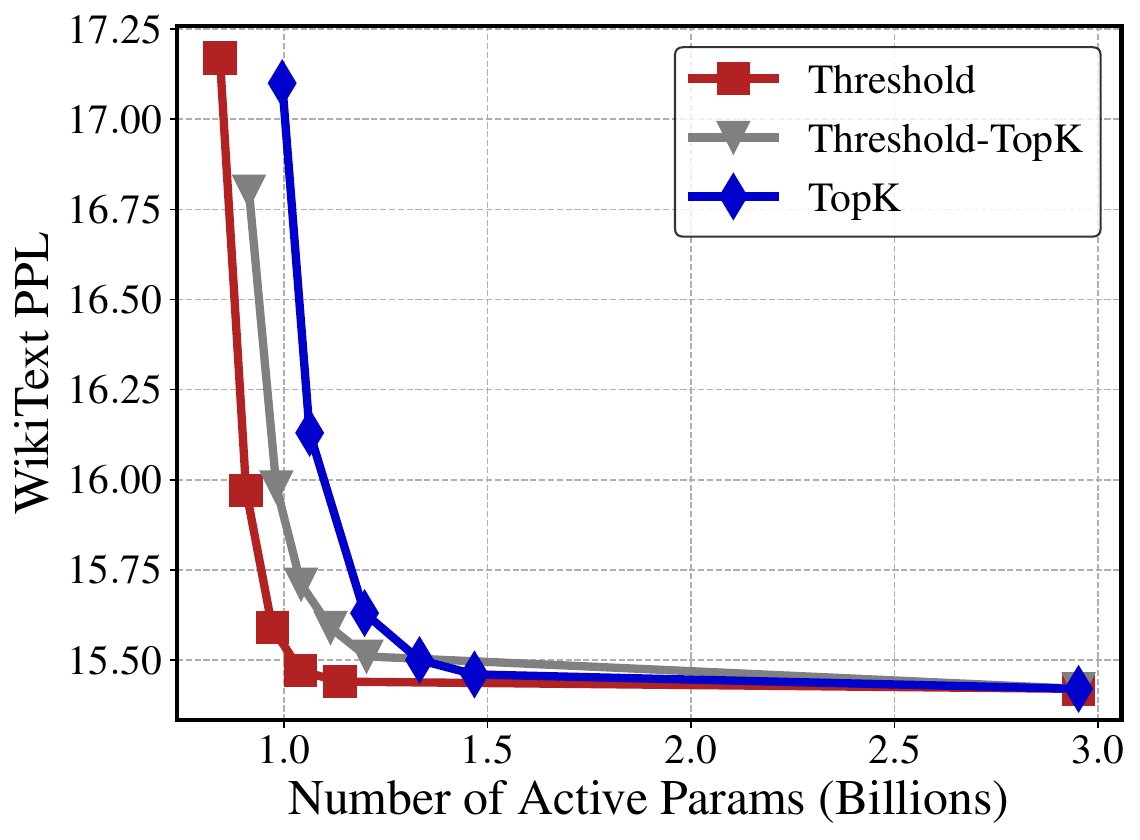}
\caption{Expert Sampling Strategy Evaluation. We assess the impact of different expert sampling strategies on the Wikitext perplexity (PPL) using our DS-MoE-3B model.}
\label{fig:exp_sample}
\end{wrapfigure}

\paragraph{Expert Sampling Strategy.} 
We explore various strategies for expert selection in our DS-MoE models. Employing a threshold on normalized expert probability yields significant reductions in active parameters but introduces challenges for real-world deployment, especially during batch inference where different tokens in the same batch may engage varying numbers of experts. To address this, we investigate two alternative sampling methods: TopK and Threshold-TopK. The TopK approach selects a set number of experts, $K$, in each MLP layer, activating all experts in the self-attention layers due to lower sparsity in self-attention. 
Meanwhile, the Threshold-TopK strategy sets a threshold for normalized expert probability, then determines the total and average number of experts activated per token in a batch, using this average as the $K$ value. These methods are demonstrated through the Wikitext perplexity and the active parameter count in our DS-MoE-3B model, as shown in Figure \ref{fig:exp_sample}. By adjusting the sparsity — either by increasing the threshold or decreasing $K$ — we find that all three expert sampling strategies strike an effective balance between computational efficiency and Wikitext perplexity. The Threshold strategy achieves the best trade-off, whereas the TopK and Threshold-TopK methods are more adaptable for real-world applications.

\begin{figure*}
\centering
\includegraphics[width=\linewidth]{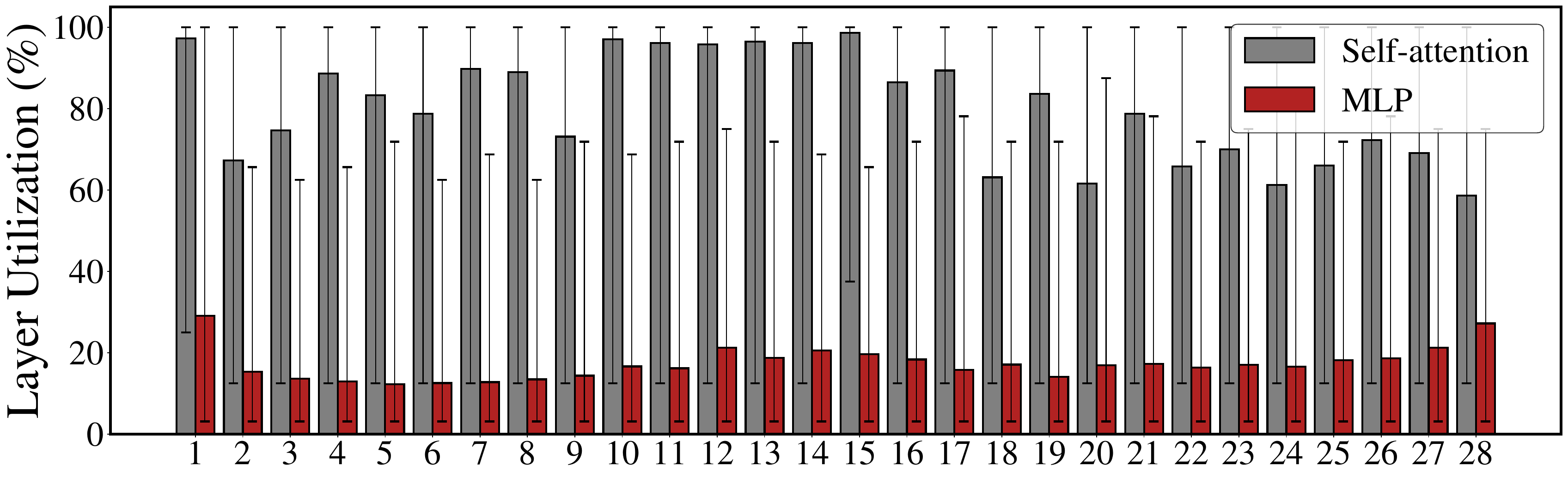}
\caption{Layer Utilization Assessment. We determine the average proportion of activated experts within both the self-attention and MLP layers. This analysis is conducted using the Wikitext dataset with our DS-MoE-3B model.}
\label{fig:layer_util}
\end{figure*}
\paragraph{Layer Utilization.} In Figure \ref{fig:layer_util}, we showcase the average percentage of active experts in each layer for a threshold value of \(\epsilon=0.48\), utilizing data from experiments conducted with the DS-MoE-3B model. The figure is augmented with error bars that depict the range of activated experts per layer, highlighting the maximum and minimum counts observed. Our findings highlight two key observations: (1) The MLP layer exhibits significantly greater sparsity compared to the self-attention layer, a trend that persists in our 6B model even when the weighting of the MI loss is identical across both the self-attention and MLP layers. (2) Within a single layer, the activated number of experts for processing different tokens exhibits substantial variance. Although sparsely trained MoEs traditionally employ a fixed number of experts, denoted as \(K\), for each layer and token, our results suggest that adhering strictly to this fixed assumption may lead to computational inefficiencies.

\subsection{GPU Inference Analysis}\label{sec:inference}
\begin{table*}[h]
    \centering
    \small
    \caption{Inference Speed of Dense Models and DS-MoE Models. Top-K represents the number of active experts in the MLP layer. We evaluate the model inference speed by measuring the latency (second) and the input token throughput (token per second). The models are deployed on HuggingFace's transformers \cite{wolf2020transformers}.}
    \begin{adjustbox}{width=\textwidth,center}
    \begin{tabular}{c|cccc|cc|cc}
        \toprule
        \multirow{2}{*}{Model} & Total & Active & \multirow{2}{*}{Top-K} & Wikitext & \multirow{2}{*}{Latency} & \multirow{2}{*}{Speedup} & \multirow{2}{*}{TPS} & \multirow{2}{*}{Speedup} \\
         & Params & Params & & PPL & & & \\
        \midrule
        Dense-3B & 2705M & 2705M & N/A & 14.77 & 4.28 & \multirow{2}{*}{1.16$\times$} &  40854.5& \multirow{2}{*}{1.51$\times$} \\
        DS-MoE-3B & 2793M & 1039M & 6 & 15.63 & 3.68 & & 61515.9 \\
        \midrule
        Dense-6B & 6186M & 6186M & N/A & 12.98 & 8.58 & \multirow{2}{*}{1.49$\times$} & 18354.2 & \multirow{2}{*}{1.91$\times$} \\
        DS-MoE-6B & 6338M & 2043M & 4 & 13.92 & 5.75 &  & 35046.7 \\
        \bottomrule
    \end{tabular}
    \label{tab:latency}
    \end{adjustbox}
\end{table*}

\begin{figure}[h]
\centering
\includegraphics[width=0.95\textwidth]{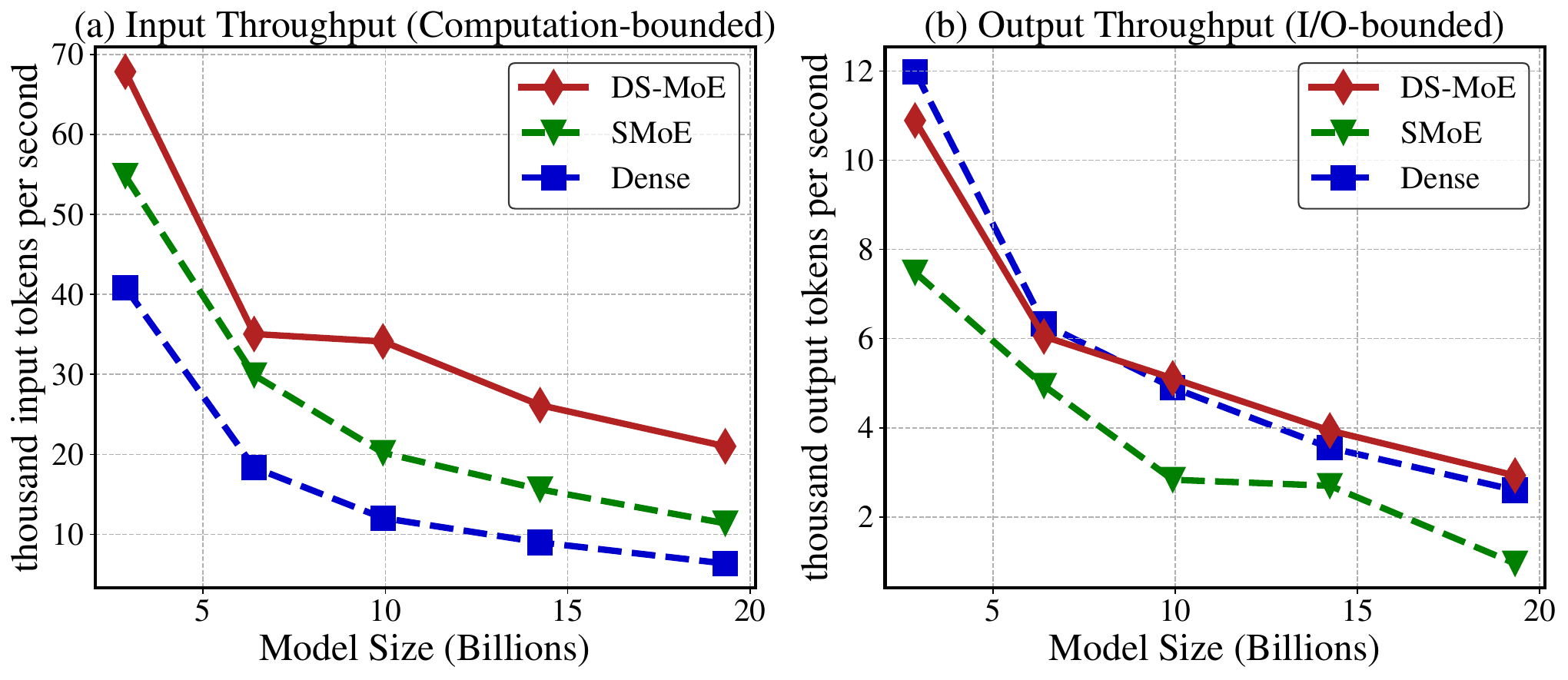}
\caption{Input and Output Token Throughput. We measure the input and output token throughput of each model on an A100-80GB GPU. The X-axis represent the model size of the dense model. In this comparison, we contrast each dense model with its corresponding performance-matched DS-MoE and SMoE models. The details of the model hyperparameters can be found in Table~\ref{tab:scaled}.}
\label{fig:speedup}
\end{figure}
In this section, we evaluate the inference performance of the dense model and our DS-MoE model on GPUs. We assess inference speed using three key metrics: (1) \textbf{Latency}, which measures the total time taken by the model to process the text input and generate a complete response. This evaluation involves processing a batch of 64 sentences, each comprising 2,000 tokens, with the model producing 20 tokens in response. (2) \textbf{Input token throughput}, which is the speed at which the model processes and encodes input tokens. For this metric, we set the input token sequence length at 256 and adjust the batch size to its maximum to optimize the utilization of GPU memory. (3) \textbf{Output token throughput}, which is measured as the model's ability to generate tokens per second from the given input tokens. This is evaluated under conditions where the model decodes 512 tokens utilizing a key-value cache mechanism. These benchmarks for assessing performance are carried out on an A100-80GB GPU. We utilize the \texttt{ParallelLinear} operation \cite{tan2024scattered} for sparse inference in the MLP layer and employ \texttt{torch.nn} \cite{paszke2019pytorch} to perform dense inference in the self-attention layer. Figure~\ref{fig:layer_util} reveals that the layer utilization of the self-attention layer consistently exceeds 60\%. We find that at this level of sparsity, sparse inference can become even slower than dense inference, primarily due to operation overheads of dynamic inference such as the duplication of intermediate tokens and the aggregation of outputs from various experts.

We first compare our DS-MoE model with the dense models regarding latency and input throughput in Table \ref{tab:latency}. We employ the TopK inference strategy, as elaborated in Section \ref{sec:abl}. As we can see in Table \ref{tab:latency}, the DS-MoE model consistently achieves a speedup across both metrics. Notably, the speedup effect amplifies on the DS-MoE-6B model with an increase in model size. This correlation is attributed to the models becoming sparser with larger parameter sizes, a phenomenon detailed in Section \ref{sec:exp-results}. Additionally, with the augmentation in model size, the models lean more towards being computation-bounded, making the operation overheads of dynamic inference increasingly insignificant. 

\begin{table}
    \centering
    \caption{Hyperparameters of the Scaled Models for the Speed Test. We maintain the total number of heads $N_\mathrm{att}\times N_\mathrm{head}$ as 32 and increase the number of layers $N_\mathrm{layer}$ to 36.}
    \begin{tabular}{c|cHcccc|cc}
        \toprule
    \multirow{2}{*}{Model} & \multirow{2}{*}{$D_{\text{emb}}$} & \multirow{2}{*}{$N_\mathrm{layer}$} & \multirow{2}{*}{$D_\mathrm{att}$} & \multirow{2}{*}{$N_\mathrm{ffd}$} & \multirow{2}{*}{$D_\mathrm{ffd}$} & \multirow{2}{*}{Top-K} & total & active\\
 & & & & & & & params & params\\
        \midrule
        Dense-10B &  \multirow{3}{*}{5120} & \multirow{3}{*}{36} & \multirow{3}{*}{160} & 1 & 20480 & 1 & 9667M & 9667M \\
        SMoE-17B &  &  & & 16 & 2560 & 2 & 17413M & 4201M \\
        DS-MoE-10B &  &  & & 32 & 640 & 4 & 9857M & 3257M \\
        \midrule
        Dense-14B & \multirow{3}{*}{6144} & \multirow{3}{*}{36} & \multirow{3}{*}{192} & 1 & 24576 & 1 & 13923M & 13923M \\
        SMoE-25B &  &  &  & 16 & 3072 & 2 & 25029M & 6004M \\
        DS-MoE-14B &  &  &  & 32 & 768 & 4 & 14149M & 4645M \\
        \midrule
        Dense-19B & \multirow{3}{*}{7168} & \multirow{3}{*}{36} & \multirow{3}{*}{224} & 1 & 28672 & 1 & 18951M & 18951M \\
        SMoE-34B &  &  &  & 16 & 3584 & 2 & 34023M & 8127M\\
        DS-MoE-19B &  &  &  & 32 & 896 & 4 & 19216M & 6278M\\
        \bottomrule
    \end{tabular}
    \label{tab:scaled}
\end{table}

Then, to thoroughly examine the inference advantages of our DS-MoE model, we conduct a comparison with both dense model and SMoE models at larger model scales. To ensure the models we compare have matched performance, we assume the SMoE to have $2\times$ the parameters in the MLP layer as both the dense and DS-MoE models, referencing Table \ref{tab:base_performance} and relevant literature \cite{dai2024deepseekmoe, shen2023moduleformer}. We enlarge the models by increasing both the embedding dimension ($D_{\text{emb}}$) and the number of hidden layers ($N_{\text{layer}}$), while keeping the number of experts and attention heads constant. The hyperparameters for these scaled models are detailed in Table \ref{tab:scaled}. We evaluate both the input and output throughput of the models, corresponding to computation-bounded and I/O-bounded scenarios, respectively. As illustrated in Figure \ref{fig:speedup}(a), in computation-bounded scenarios, our DS-MoE model demonstrates a significantly higher input throughput, particularly when compared to the dense model, showcasing its computational efficiency. In contrast, Figure \ref{fig:speedup}(b) reveals that while our DS-MoE model achieves comparable throughput to the dense model, it significantly outperforms the SMoE model in terms of throughput, highlighting its parameter efficiency.

\paragraph{Comparison with other MoEs.} We further deploy our DS-MoE models with vLLM \cite{kwon2023efficient} to benchmark our inference speed against other models at the 7B performance tier. For comparison, we select the Mistral-7B \cite{jiang2023mistral}, which stands out as one of the leading open-source 7B models. According to Table \ref{tab:speed-comparison}, our DS-MoE-6B model demonstrates a speed increase of $1.86\times$ and $1.64\times$ over the Mistral-7B on A100-80GB GPU and H100-80GB GPU, respectively. For MoEs, we choose DeepSeekMoE-16B \cite{dai2024deepseekmoe} and Qwen1.5-MoE-A2.7B \cite{qwen}. Both of them are sparsely trained and comparable to the performance of 7B dense models. Table \ref{tab:speed-comparison} illustrates that DeepSeekMoE-16B and Qwen1.5-MoE-A2.7B possess active parameters similar to those of DS-MoE-6B, but their total weights nevertheless occupy more than $2\times$ the GPU memory compared to DS-MoE-6B. This affects both the maximum batch size that a GPU can handle and the I/O latency, subsequently impacting throughput. As shown in Table \ref{tab:speed-comparison}, our DS-MoE-6B model is $1.50\times$ and $1.27\times$ faster than Qwen1.5-MoE-A2.7B on A100-80GB GPU and H100-80GB GPU, respectively.
\begin{table*}[h]
    \centering
    \small
    \caption{Speed Comparison with other MoEs. We deploy our Dense-6B and DS-MoE-6B models with vLLM \cite{kwon2023efficient} and test the performance under the experimental setup where the number of input tokens are 1000 and output tokens are 1000. We measure the performance with two metrics: (1) \textbf{Throughput}: requests processed per second; (2) \textbf{TPS}: tokens processed per second. The GPU memory utilization is set to be 0.9.}
    \begin{tabular}{c|ccc|cc|cc}
        \toprule
        \multirow{3}{*}{Model} & \multirow{2}{*}{Total} & \multirow{2}{*}{Active} & \multirow{2}{*}{Model} & \multicolumn{2}{c|}{\multirow{2}{*}{\textbf{A100-80GB}}} & \multicolumn{2}{c}{\multirow{2}{*}{\textbf{H100-80GB}}} \\
         & \multirow{2}{*}{Params} & \multirow{2}{*}{Params} & \multirow{2}{*}{Memory} & \multirow{2}{*}{Throughput} & \multirow{2}{*}{TPS} & \multirow{2}{*}{Throughput} & \multirow{2}{*}{TPS}\\
         & & & & & & & \\
        \midrule
        Dense-6B & 6.4B & 6.4B & 12.3 GiB & 1.04 & 2079.8 & 1.40 & 2808.7 \\
        Mistral-7B & 7.2B & 7.2B & 13.5 GiB & 1.07 & 2140.8 & 1.52 & 3047.4 \\
        DeepSeekMoE & 17.3B & 2.8B & 30.5 GiB & 1.17 & 2330.1 & 1.57 & 3144.1 \\
        Qwen1.5-MoE & 16.4B & 2.7B & 26.7 GiB & 1.33 & 2665.7 & 1.81 & 3616.9 \\
        DS-MoE-6B & 6.5B & 2.2B & 12.6 GiB & 2.00 & 3992.8 & 2.30 & 4603.9 \\
        \bottomrule
    \end{tabular}
    \label{tab:speed-comparison}
\end{table*}

\section{Conclusion}
In this study, we investigate the application of dense pre-training, sparse inference for MoE (DS-MoE) models. Our analysis reveals that traditional sparse training methods do not optimize MoE models efficiently during the training phase, leading to suboptimal parameter effectiveness in the resulting MoE models. In contrast, we adopt a strategy where gradients are densely propagated to both the router and experts. This approach ensures our MoE models maintain the parameter efficiency comparable to dense models, while the computation efficiency similar with MoE models. Moreover, we demonstrate that our DS-MoE model achieves superior inference throughput in both computation-bounded and I/O-bounded scenarios. Upon comparing our DS-MoE against other MoE models, it is evident that our DS-MoE model achieves the highest throughput.

\bibliography{colm2024_conference}
\bibliographystyle{colm2024_conference}

\end{document}